\def\year{2019}\relax
\documentclass[letterpaper]{article} %DO NOT CHANGE THIS
\usepackage{aaai19}  %Required
\usepackage{times}  %Required
\usepackage{helvet}  %Required
\usepackage{courier}  %Required
\usepackage{url}  %Required
\usepackage{graphicx}  %Required
\usepackage{amsmath}
\usepackage{amsthm}
\usepackage{amssymb}
\DeclareMathOperator*{\E}{\mathbb{E}}

\usepackage{algorithmicx}
\usepackage[noend]{algpseudocode}
\usepackage{algorithm}
\usepackage{booktabs}

\DeclareMathOperator*{\argmax}{\arg\max}
\newtheorem{theorem}{Theorem}

\begin{document}
	% The file aaai.sty is the style file for AAAI Press 
	% proceedings, working notes, and technical reports.
	%
	\title{Melding the Data-Decisions Pipeline: Decision-Focused Learning for Combinatorial Optimization}
	\author{Bryan Wilder,  Bistra Dilkina, Milind Tambe\\Center for Artificial Intelligence in Society, University of Southern California\\\{bwilder, dilkina, tambe\}@usc.edu
	}
	\maketitle
	\begin{abstract}
		Creating impact in real-world settings requires artificial intelligence techniques to span the full pipeline from data, to predictive models, to decisions. These components are typically approached separately: a machine learning model is first trained via a measure of predictive accuracy, and then its predictions are used as input into an optimization algorithm which produces a decision. However, the loss function used to train the model may easily be misaligned with the end goal, which is to make the best decisions possible. Hand-tuning the loss function to align with optimization is a difficult and error-prone process (which is often skipped entirely). 
		
		We focus on combinatorial optimization problems and introduce a general framework for decision-focused learning, where the machine learning model is directly trained in conjunction with the optimization algorithm to produce high-quality decisions. Technically, our contribution is a means of integrating common classes of discrete optimization problems into deep learning or other predictive models, which are typically trained via gradient descent. The main idea is to use a continuous relaxation of the discrete problem to propagate gradients through the optimization procedure. We instantiate this framework for two broad classes of combinatorial problems: linear programs and submodular maximization. Experimental results across a variety of domains show that decision-focused learning often leads to improved optimization performance compared to traditional methods. We find that standard measures of accuracy are not a reliable proxy for a predictive model's utility in optimization, and our method's ability to specify the true goal as the model's training objective yields substantial dividends across a range of decision problems.
		
	\end{abstract}

	\section{Introduction}
	
	The goal in many real-world applications of artificial intelligence is to create a pipeline from data, to predictive models, to decisions. Together, these steps enable a form of evidence-based decision making which has transformative potential across domains such as healthcare, scientific discovery, transportation, and more \cite{horvitz2010data,horvitz2010healthcare}. This pipeline requires two technical components: machine learning models and optimization algorithms. Machine learning models use the data to predict unknown quantities; optimization algorithms use these predictions to arrive at a decision which maximizes some objective. Our concern here is combinatorial optimization, which is ubiquitous in real-world applications of artificial intelligence, ranging from matching applicants to public housing to selecting a subset of movies to recommend. We focus on common classes of combinatorial problems which have well-structured continuous relaxations, e.g., linear programs and submodular maximization. A vast literature has been devoted to combinatorial optimization \cite{korte2012combinatorial}. Importantly though, optimization is often insufficient without the broader pipeline because the objective function is unknown and must predicted via machine learning. 
	
	While machine learning has witnessed incredible growth in recent years, the two pieces of the pipeline are treated entirely separately by typical training approaches. That is, a system designer will first train a predictive model using some standard measure of accuracy, e.g., mean squared error for a regression problem. Then, the model's predictions are given as input to the optimization algorithm to produce a decision. Such \emph{two-stage} approaches are extremely common across many domains \cite{wang2006cope,fang2016deploying,mukhopadhyay2017prioritized,xue2016avicaching}. This process is justified when the predictive model is perfect, or near-so, since completely accurate predictions also produce the best decisions. However, in complex learning tasks, all models will make errors and the training process implicitly trades off where these errors will occur. When prediction and optimization are separate, this tradeoff is divorced from the goal of the broader pipeline: to make the best decision possible. 
	
	We propose a \emph{decision-focused learning} framework which melds the data-decisions pipeline by integrating prediction and optimization into a single end-to-end system. That is, the predictive model is trained using the quality of the decisions which it induces via the optimization algorithm. Similar ideas have recently been explored in the context of convex optimization \cite{donti2017task}, but to our knowledge ours is the first attempt to train machine learning systems for performance on \emph{combinatorial} decision-making problems. Combinatorial settings raise new technical challenges because the optimization problem is discrete. However, machine learning systems (e.g., deep neural networks) are often trained via gradient descent. 
	
	Our first contribution is a general framework for training machine learning models via their performance on combinatorial problems. The starting point is to relax the combinatorial problem to a continuous one. Then, we analytically differentiate the optimal solution to the continuous problem as a function of the model's predictions. This allows us to train using a continuous proxy for the discrete problem. At test time, we round the continuous solution to a discrete point. 
	
	Our second contribution is to instantiate this framework for two broad classes of combinatorial problems: linear programs and submodular maximization problems. Linear programming encapsulates a number of classical problems such as shortest path, maximum flow, and bipartite matching. Submodular maximization, which reflects the intuitive phenomena of diminishing returns, is also ubiquitous; applications range from social networks \cite{kempe_maximizing_2003} to recommendation systems \cite{viappiani2010optimal}. In each case, we resolve a set of technical challenges to produce well-structured relaxations which can be efficiently differentiated through. 
	
	Finally, we give an extensive empirical investigation, comparing decision-focused and traditional methods on a series of domains. Decision-focused methods often improve performance for the pipeline as a whole (i.e., decision quality) despite worse predictive accuracy according to standard measures. Intuitively, the predictive models trained via our approach focus specifically on qualities which are important for making good decisions. By contrast, more generic methods produce predictions where error is distributed in ways which are not aligned with the underlying task.
	
	\section{Problem description}
	
	We consider combinatorial optimization problems of the form $\max_{x\in \mathcal{X}} f(x, \theta)$, where $\mathcal{X}$ is a discrete set enumerating the feasible decisions. Without loss of generality, $\mathcal{X} \subseteq \{0,1\}^n$ and the decision variable $x$ is a binary vector. The objective $f$ depends on a parameter $\theta \in \Theta$. If $\theta$ were known exactly, a wide range of existing techniques could be used to solve the problem. In this paper, we consider the challenging (but prevalent) case where $\theta$ is unknown and must be inferred from data. For instance, in bipartite matching, $x$ represents whether each pair of nodes were matched and $\theta$ contains the reward for matching each pair. In many applications, these affinities are learned from historical data.
	
	Specifically, the decision maker observes a feature vector $y \in \mathcal{Y}$ which is correlated with $\theta$. This introduces a learning problem which must be solved prior to optimization. As in classical supervised learning, we formally model $y$ and $\theta$ as drawn from a joint distribution $P$. Our algorithm will observe training instances $(y_1, \theta_1)...(y_N, \theta_N)$ drawn iid from $P$. At test time, we are give a feature vector $y$ corresponding to an \emph{unobserved} $\theta$. Our algorithm will use $y$ to predict a parameter value $\hat{\theta}$. Then, we will solve the optimization problem $\max_x f(x, \hat{\theta})$ to obtain a decision $x^*$. Our utility is the objective value that $x^*$ obtains with respect to the \emph{true but unknown} parameter $\theta$, $f(x^*, \theta)$. 
	
	Let $m : \mathcal{Y} \to \Theta$ denote a model mapping observed features to parameters. Our goal is to (using the training data) find a model $m$ which maximizes expected performance on the underlying optimization task. Define $x^*(\theta) = \arg\max_{x \in \mathcal{X}} f(x, \theta)$ to be the optimal $x$ for a given $\theta$. The end goal of the data-decisions pipeline is to maximize
	
	\begin{align} \label{eq:task-objective}
	\E_{y, \theta \sim P}\left[f(x^*(m(y)), \theta)\right]
	\end{align}
	
	The classical approach to this problem is a \emph{two-stage} method which first learns a model using a task-agnostic loss function (e.g., mean squared error) and then uses the learned model to solve the optimization problem. The model class will have its own parameterization, which we denote by $m(y, \omega)$. For instance, the model class could consist of deep neural networks where $\omega$ denotes the weights. The two-stage approach first solves the problem $\min_\omega \E_{y, \theta \sim P}\left[\mathcal{L}(\theta, m(y, \omega))\right]$, where $\mathcal{L}$ is a loss function. Such a loss function measures the overall ``accuracy" of the model's predictions but does not specifically consider how $m$ will fare when used for decision making. The question we address is whether it is possible to do better by specifically training the model to perform well on the decision problem. 
	
	\section{Previous work}
	
 	There is a growing body of research at the interface of machine learning and discrete optimization \cite{vinyals2015pointer,bertsimas2017optimal,khalil2017tree,khalil2017graphs}. However, previous work largely focuses on either using discrete optimization to find an accuracy-maximizing predictive model or using machine learning to speed up optimization algorithms. Here, we pursue a deeper synthesis; to our knowledge, this work is the first to train predictive models using combinatorial optimization performance with the goal of improving decision making.
	
	The closest work to ours in motivation is \cite{donti2017task}, who study task-based convex optimization. Their aim is to optimize a convex function which depends on a learned parameter. As in their work, we use the idea of differentiating through the KKT conditions. However, their focus is entirely on continuous problems. Our discrete setting raises new technical challenges, highlighted below. Elmachtoub and Grigas \shortcite{elmachtoub2017smart} also propose a means of integrating prediction and optimization; however, their method applies strictly to linear optimization and focuses on linear predictive models while our framework applies to nonlinear problems with more general models (e.g., neural networks). Finally, some work has noted that two-stage methods lead to poor optimization performance in specific domains \cite{beygelzimer2009offset,ford2015beware}. 
	
	Our work is also related to recent research in structured prediction \cite{belanger2017end,tu2018learning,niculae2018sparsemap,djolonga2017differentiable}. which aims to make a prediction lying in a discrete set. This is fundamentally different than our setting since their goal is to \emph{predict} an external quantity, not to \emph{optimize} and find the best decision possible. However, structured prediction sometimes integrates a discrete optimization problem as a module within a larger neural network. The closest such work technically to ours is \cite{tschiatschek2018differentiable}, who design a differentiable algorithm for submodular maximization in order to predict choices made by users. Their approach is to introduce noise into the standard greedy algorithm, making the probability of outputting a given set differentiable. There are two key differences between our approaches. First, their approach does not apply to the decision-focused setting because it maximizes the likelihood of a \emph{fixed} set but cannot optimize for finding the best set. Second, exactly computing gradients for their algorithm requires marginalizing over the $k!$ possible permutations of the items, forcing a heuristic approximation to the gradient. Our approach allows closed-form differentiation. 
	
	Some deep learning architectures differentiate through gradient descent steps, related to our approach in the submodular setting. Typically, previous approaches explicitly unroll  $T$ iterations of gradient descent in the computational graph \cite{domke2012generic}. However, this approach is usually employed for \emph{unconstrained} problems where each iteration is a simple gradient step. By contrast, our combinatorial problems are constrained, requiring a projection step to enforce feasibility. Unrolling the projection step may be difficult, and would incur a large computational cost. We instead exploit the fact that gradient ascent converges to a local optimum and analytically differentiate via the KKT conditions.

	\section{General framework}
	
	Our goal is to integrate combinatorial optimization into the loop of gradient-based training. That is, we aim to directly train the predictive model $m$ by running gradient steps on the objective in Equation \ref{eq:task-objective}, which integrates both prediction and optimization. The immediate difficulty is the dependence on $x^*(m(y, \omega))$. This term is problematic for two reasons. First, it is a discrete quantity since $x^*$ is a decision from a binary set. This immediately renders the output nondifferentiable with respect to the model parameters $\omega$.  Second, even if $x^*$ were continuous, it is still defined as the solution to an optimization problem, so calculating a gradient requires us to differentiate through the argmax operation.
	
	We resolve both difficulties by considering a continuous relaxation of the combinatorial decision problem. We show that for a broad class of combinatorial problems, there are appropriate continuous relaxations such that we can analytically obtain derivatives of the continuous optimizer with respect to the model parameters. This allows us to train any differentiable predictive model via gradient descent on a continuous surrogate to Equation \ref{eq:task-objective}. At test time, we solve the true discrete problem by rounding the continuous point.  
	
	More specifically, we relax the discrete constraint $x \in \mathcal{X}$ to the continuous one $x \in conv(\mathcal{X})$ where $conv$ denotes the convex hull. Let $x(\theta) = \arg\max_{x \in conv(\mathcal{X})}f(x, \theta)$ denote the optimal solution to the continuous problem. To train our predictive model, we would like to compute gradients of the whole-pipeline objective given by Equation \ref{eq:task-objective}, replacing the discrete quantity $x^*$ with the continuous $x$. We can obtain a stochastic gradient estimate by sampling a single $(y, \theta)$ from the training data. On this sample, the chain rule gives
	
	\begin{align*}
	\frac{d f (x(\hat{\theta}), \theta)}{d \omega} = \frac{d f(x(\hat{\theta}), \theta)}{d x(\hat{\theta})} \frac{d x(\hat{\theta})}{d \hat{\theta}} \frac{d \hat{\theta}}{d \omega}
	\end{align*}

	The first term is just the gradient of the objective with respect to the decision variable $x$, and the last term is the gradient of the model's predictions with respect to its own internal parameterization. 
	
	The key is computing the middle term, which measures how the optimal decision changes with respect to the prediction $\hat{\theta}$. For continuous problems, the optimal continuous decision $x$ must satisfy the KKT conditions (which are sufficient for convex problems). The KKT conditions define a system of linear equations based on the gradients of the objective and constraints around the optimal point. Is is known that by applying the implicit function theorem, we can differentiate the solution to this linear system \cite{gould2016differentiating,donti2017task}. In more detail, recall that our continuous problem is over $conv(\mathcal{X})$, the convex hull of the discrete feasible solutions. This set is a polytope, which can be represented via linear equalities as the set $\{x: Ax\leq b\}$ for some matrix $A$ and vector $b$. Let $(x, \lambda)$ be pair of primal and dual variables which satisfy the KKT conditions. Then differentiating the conditions yields that
	
	\begin{align}\label{eq:kkt}
	\begin{bmatrix}
	&\nabla^2_x f(x, \theta) & A^T\\
	& diag(\lambda) A & diag(Ax-b)\\
	\end{bmatrix}
	\begin{bmatrix}
	\frac{d x}{d\theta}\\
	\frac{d \lambda}{d\theta}
	\end{bmatrix}
	=
	\begin{bmatrix}
	\frac{d \nabla_x f(x, \theta)}{d\theta} \\
	0
	\end{bmatrix}
	\end{align} 
	
	By solving this system of linear equations, we can obtain the desired term $\frac{dx}{d\theta}$. However, the above approach is a general framework; our main technical contribution is to instantiate it for specific classes of combinatorial problems. Specifically, we need (1) an appropriate continuous relaxation, along with a means of solving the continuous optimization problem and (2) efficient access to the terms in Equation \ref{eq:kkt} which are needed for the backward pass (i.e., gradient computation). We provide both ingredients for two broad classes of problems: linear programming and submodular maximization. In each setting, the high-level challenge is to ensure that the continuous relaxation is differentiable, a feature not satisfied by naive alternatives. We also show how to efficiently compute terms needed for the backward pass, especially for the more intricate submodular case.

	\subsection{Linear programming}
	
	The first setting that we consider is combinatorial problems which can be expressed as a linear program with equality and inequality constraints in the form
	
	\begin{align}
	&\max \theta^T x  \,\,\,  \text{s.t. }\,Ax = b, \,\, Gx \leq h \label{problem:lp}
	\end{align}
	
	Example problems include shortest path, maximum flow, bipartite matching, and a range of other domains. For instance, in a shortest path problem $\theta$ contains the cost for traversing each edge, and we are interested in problems where the true costs are unknown and must be predicted. Since the LP can be regarded as a continuous problem (it just happens that the optimal solutions in these example domains are integral), we could attempt to apply Equation \ref{eq:kkt} and differentiate the solution. This approach runs into an immediate difficulty: the optimal solution to an LP may not be differentiable (or even continuous) with respect to $\theta$. This is because the optimal solution may ``jump" to a different vertex. Formally, the left-hand side matrix in Equation \ref{eq:kkt} becomes singular since $\nabla_x^2 f(x, \theta)$ is always zero.  
	We resolve this challenge by instead solving the regularized problem
	
	\begin{align} 
	&\max \theta^T x - \gamma ||x||_2^2 \,\,\,  \text{s.t. }\,Ax = b, \,\, Gx \leq h\label{eq:lp-quad}
	\end{align}
	
	which introduces a penalty proportional to the squared norm of the decision vector. This transforms the LP into a strongly concave quadratic program (QP). The Hessian is given by $\nabla_x^2 f(x, \theta) = -2\gamma I$ (where $I$ is the identity matrix), which renders the solution differentiable under mild conditions (see supplement for proof):
	
	\begin{theorem}
		Let $x(\theta)$ denote the optimal solution of Problem \ref{eq:lp-quad}. Provided that the problem is feasible and all rows of $A$ are linearly independent, $x(\theta)$ is differentiable with respect to $\theta$ almost everywhere. If $A$ has linearly dependent rows, removing these rows yields an equivalent problem which is differentiable almost everywhere. Wherever $x(\theta)$ is differentiable, it satisfies the conditions in Equation \ref{eq:kkt}. 
	\end{theorem}
	
	Moreover, we can control the loss that regularization can cause on the original, linear problem:
	
	\begin{theorem}
		Define $D = \max_{x, y\in conv(\mathcal{X})} ||x - y||^2$ as the squared diameter of the feasible set and $OPT$ to be the optimal value for Problem \ref{problem:lp}. We have $\theta^\top x(\theta) \geq OPT - \gamma D$.
	\end{theorem}
	
	Together, these results give us a differentiable surrogate which still enjoys an approximation guarantee relative to the integral problem.  Computing the backward pass via Equation \ref{eq:kkt} is now straightforward since all the relevant terms are easily available. Since $\nabla_x \theta^\top x = \theta$, we have $\frac{d \nabla_x f(x, \theta)}{d\theta} = I$. All other terms are easily computed from the optimal primal-dual pair $(x, \lambda)$ which is output by standard QP solvers. We can also leverage a recent QP solver \cite{amos2017optnet} which maintains a factorization of the KKT matrix for a faster backward pass. At test time, we simply set $\gamma = 0$ to produce an integral decision.
	
	\subsection{Submodular maximization}
	We consider problems where the underlying objective to maximize a set function $f: 2^V \to R$, where $V$ is a ground set of items. A set function is \emph{submodular} if for any $A \subseteq B$ and any $v \in V\setminus B$, $f(A \cup \{v\}) - f(A) \geq f(B \cup \{v\}) - f(B)$. We will restrict our consideration to submodular functions which are \emph{monotone} ($f(A \cup \{v\}) - f(A) \geq 0 \,\, \forall A, v$) and \emph{normalized} $f(\emptyset)  = 0$. This class of functions contains many combinatorial problems which have been considered in machine learning and artificial intelligence (e.g., influence maximization, facility location, diverse subset selection, etc.). We focus on the cardinality-constrained optimization problem $\max_{|S| \leq k} f(S)$, though our framework easily accommodates more general matroid constraints.

	\textbf{Continuous relaxation: } We employ the canonical continuous relaxation for submodular set functions, which associates each set function $f$ with its \emph{multilinear extension} $F$ \cite{calinescu2011maximizing}. We can view a set function as defined on the domain $\{0,1\}^{|V|}$, where each element is an indicator vector which the items contained in the set. The extension $F$ is a continuous function defined on the hypercube $[0,1]^{|V|}$. We interpret a given fraction vector $x \in [0,1]^{|V|}$ as giving the marginal probability that each item is included in the set. $F(x)$ is the expected value of $f(S)$ when each item $i$ is included in $S$ independently with probability $x_i$. In other words, $F(x) = \sum_{S \subseteq V} f(S) \prod_{i\in S}x_i \prod_{i\not\in S} 1-x_i$. While this definition sums over exponentially many terms, arbitrarily close approximations can be obtained via random sampling. Further, closed forms are available for many cases of interest \cite{iyer2014monotone}. Importantly, well-known rounding algorithms \cite{calinescu2011maximizing} can convert a fractional point $x$ to a set $S$ satisfying $\E[f(S)] \geq F(x)$; i.e., the rounding is lossless. 
	
	As a proxy for the discrete problem $\max_{|S| \leq k} f(S)$, we can instead solve $\max_{x \in conv(\mathcal{X})} F(x)$, where $\mathcal{X} = \{x \in \{0,1\}^{|V|}: \sum_i x_i \leq k\}$. Unfortunately, $F$ is not in general concave. Nevertheless, many first-order algorithms still obtain a constant factor approximation. For instance, a variant of the Frank-Wolfe algorithm solves the continuous maximization problem with the optimal approximation ratio of $(1 - 1/e)$ \cite{calinescu2011maximizing,bian2017guaranteed}. 
	
	However, non-concavity complicates the problem of differentiating through the continuous optimization problem. Any polynomial-time algorithm can only be guaranteed to output a \emph{local} optimum, which need not be unique (compared to strongly convex problems, where there is a single global optimum). Consequently, the algorithm used to select $x(\theta)$ might return a \emph{different} local optimum under an infinitesimal change to $\theta$. For instance, the Frank-Wolfe algorithm (the most common algorithm for continuous submodular maximization) solves a linear optimization problem at each step. Since (as noted above), the solution to a linear problem may be discontinuous in $\theta$, this could render the output of the optimization problem nondifferentiable.
	
	We resolve this difficulty through a careful choice of optimization algorithm for the forward pass. Specifically, we use apply projected stochastic gradient ascent (SGA), which has recently been shown to obtain a $\frac{1}{2}$-approximation for continuous submodular maximization \cite{Hassani2017gradient}. Although SGA is only guaranteed to find a local optimum, each iteration applies purely differentiable computations (a gradient step and projection onto the set $conv(\mathcal{X})$), and so the final output after $T$ iterations will be differentiable as well. Provided that $T$ is sufficiently large, this output will converge to a local optimum, which must satisfy the KKT conditions. Hence, we can apply our general approach to the local optimum returned by SGA.  The following theorem shows that the local optima of the multilinear extension are differentiable:
	
	\begin{theorem} \label{theorem:submod-diff}
		Suppose that $x^*$ is a local maximum of the multilinear extension, i.e,., $\nabla_x F(x^*, \theta) = 0$ and $\nabla_x^2 F(x^*, \theta) \succ 0$. Then, there exists a neighborhood $\mathcal{I}$ around $x^*$ such that the maximizer of $F(\cdot, \theta)$ within $\mathcal{I} \cap conv(\mathcal{X})$ is differentiable almost everywhere as a function of $\theta$, with $\frac{d x(\theta)}{d\theta}$ satisfying the conditions in Equation \ref{eq:kkt}. 
	\end{theorem}

	We remark that Theorem \ref{theorem:submod-diff} requires a local maximum, while gradient ascent may in theory find saddle points. However, recent work shows that random perturbations ensure that gradient ascent quickly escapes saddle points and finds an approximate local optimum \cite{jin2017escape}. 
	
	\textbf{Efficient backward pass:} We now show how the terms needed to compute gradients via Equation \ref{eq:kkt} can be efficiently obtained. In particular, we need access to the optimal dual variable $\lambda$ as well as the term $\frac{d \nabla_x F(x, \theta)}{d\theta}$. These were easy to obtain in the LP setting but the submodular setting requires some additional analysis. Nevertheless, we show that both can be obtained efficiently. 
	
	\textbf{Optimal dual variables:} SGA only produces the optimal primal variable $x$, not the corresponding dual variable $\lambda$ which is required to solve Equation \ref{eq:kkt} in the backward pass. We show that for cardinality-constrained problems, we can obtain the optimal dual variables analytically given a primal solution $x$. Let $\lambda_i^L$ be the dual variable associated with the constraint $x_i \geq 0$, $\lambda_i^U$ with $x_i \leq 1$ and $\lambda^S$ with $\sum_i x_i \leq k$. By differentiating the Lagrangian, any optimum satisfies
	
	\begin{align*}
	\nabla_{x_i} f(x) - \lambda_i^L + \lambda_i^U + \lambda_i^S = 0 \quad \forall i
	\end{align*}
	
	where complementary slackness requires that $\lambda_i^L = 0$ if $x_i > 0$ and $\lambda_i^U = 0$ if $x_i < 1$. Further, it is easy to see that for all $i$ with $0 < x_i < 1$, $\nabla_{x_i} f(x)$ must be equal. Otherwise, $x$ could not be (locally) optimal since we could increase the objective by finding a pair $i,j$ with $\nabla_{x_i} f(x) > \nabla_{x_j} f(x)$, increasing $x_i$, and decreasing $x_j$. Let $\nabla_{*}$ denote the shared gradient value for fractional entries. We can solve the above equation and express the optimal dual variables as
	
	\begin{align*}
	\lambda^S = -\nabla_*, \,\,\,\, \lambda_i^L = \lambda^S - \nabla_{x_i} f, \,\,\,\, \lambda_i^U = \nabla_{x_i} f - \lambda^S
	\end{align*}
	
	where the expressions for $\lambda_i^L$ and $\lambda_i^U$ apply only when $x_i = 0$ and $x_i = 1$ respectively (otherwise, complementary slackness requires these variables be set to 0). 
	
	\textbf{Computing $\mathbf{\frac{d}{d\theta} \nabla_x F(x, \theta)}$: } We show that this term can be obtained in closed form for the case of probabilistic coverage functions, which includes many cases of practical interest (e.g.\ budget allocation, sensor placement, facility location, etc.). However, our framework can be applied to arbitrary submodular functions; we focus here on coverage functions just because they are particularly common in applications. A coverage function takes the following form. There a set of items $U$, and each $j \in U$ has a weight $w_j$. The algorithm can choose from a ground set $V$ of actions. Each action $a_i$ covers each item $j$ independently with probability $\theta_{ij}$. We consider the case where the probabilities $\theta$ are be unknown and must be predicted from data. For such problems, the multilinear extension has a closed form 
	
	\begin{align*}
	F(x, \theta) = \sum_{j \in U} w_j \left(1 - \prod_{i \in V} 1 - x_{ij}\theta_{ij}\right)
	\end{align*}
	
	and we can obtain the expression
	
	\begin{align*}
	\frac{d}{d\theta_{kj}} \nabla_{x_i} F(x, \theta) = \begin{cases}
	- \theta_{ij} x_k \prod_{\ell \not= i,k} 1 - x_\ell \theta_{\ell j} & \text{if } k \not= i\\
	\prod_{k \not=i} 1 - x_k \theta_{kj} & \text{otherwise}.
	\end{cases}
	\end{align*}

	\section{Experiments}
	
	We conduct experiments across a variety of domains in order to compare our decision-focused learning approach with traditional two stage methods. We start out by describing the experimental setup for each domain. Then, we present results for the complete data-decisions pipeline in each domain (i.e., the final solution quality each method produces on the optimization problem). We find that decision-focused learning almost always outperforms two stage approaches. To investigate this phenomenon, we show more detailed results about what each model learns. Two stage approaches typically learn predictive models which are more accurate according to standard measures of machine learning accuracy. However, decision-focused methods learn qualities which are important for optimization performance even if this leads to lower accuracy in an overall sense. 
	
	\begin{table*}\centering
		%		\ra{1.3}
		\fontsize{9}{9}\selectfont
		\caption{Solution quality of each method for the full data-decisions pipeline.}\label{table:opt}
		\begin{tabular}{@{}r|ccccccccc@{}}\toprule
			& \multicolumn{3}{c}{Budget allocation} & \phantom{ab}& \multicolumn{1}{c}{ Matching} &
			\phantom{ac} & \multicolumn{3}{c}{Diverse recommendation}\\
			\cmidrule{2-4} \cmidrule{6-6} \cmidrule{8-10} 
			$k = $ & $5$ & $10$ & $20$ && $-$  && $5$ & $10$ & $20$\\ \midrule
NN1-Decision & \textbf{49.18 $\pm$ 0.24} & \textbf{72.62 $\pm$ 0.33} & \textbf{98.95 $\pm$ 0.46} && 2.50 $\pm$ 0.56 && \textbf{15.81 $\pm$ 0.50} & \textbf{29.81 $\pm$ 0.85} & \textbf{52.43 $\pm$ 1.23}\\
NN2-Decision & 44.35 $\pm$ 0.56 & 67.64 $\pm$ 0.62 & 93.59 $\pm$ 0.77 && \textbf{6.15 $\pm$ 0.38} && 13.34 $\pm$ 0.77 & 26.32 $\pm$ 1.38 & 47.79 $\pm$ 1.96\\
NN1-2Stage & 32.13 $\pm$ 2.47 & 45.63 $\pm$ 3.76 & 61.88 $\pm$ 4.10 && 2.99 $\pm$ 0.76 && 4.08 $\pm$ 0.16 & 8.42 $\pm$ 0.29 & 19.16 $\pm$ 0.57\\
NN2-2Stage & 9.69 $\pm$ 0.05 & 18.93 $\pm$ 0.10 & 36.16 $\pm$ 0.18 && 3.49 $\pm$ 0.32 && 11.63 $\pm$ 0.43 & 22.79 $\pm$ 0.66 & 42.37 $\pm$ 1.02\\
RF-2Stage & \textbf{48.81 $\pm$ 0.32} & \textbf{72.40 $\pm$ 0.43} & \textbf{98.82 $\pm$ 0.63} && 3.66 $\pm$ 0.26 && 7.71 $\pm$ 0.18 & 15.73 $\pm$ 0.34 & 31.25 $\pm$ 0.64\\
Random & 9.69 $\pm$ 0.04 & 18.92 $\pm$ 0.09 & 36.13 $\pm$ 0.14 && 2.45 $\pm$ 0.64 && 8.19 $\pm$ 0.19 & 16.15 $\pm$ 0.35 & 31.68 $\pm$ 0.71\\
			\bottomrule
		\end{tabular}
	\end{table*}
	
	\textbf{Budget allocation: }We start with a synthetic domain which allows us to illustrate how our methods differ from traditional approaches and explore when improved decision making is achievable. This example concerns budget allocation, a submodular maximization problem which models an advertiser's choice of how to divide a finite budget $k$ between a set of channels. There is a set of customers $R$ and the objective is $f(S) = \sum_{v \in R} 1 - \prod_{u \in S} (1 - \theta_{uv})$, where $\theta_{uv}$ is the probability that advertising on channel $u$ will reach customer $v$. This is the expected number of customers reached. Variants on this problem have been the subject of a great deal of research \cite{alon2012optimizing,soma2014optimal,miyauchi2015threshold}.  
	
	In our problem, the matrix $\theta$ is not known in advance and must be learned from data. The ground truth matrices were generated using the Yahoo webscope \cite{yahoowebscope} dataset which logs bids placed by advertisers on a set of phrases. In our problem, the phrases are channels and the accounts are customers. Each instance samples a random subset of 100 channels and 500 customers. For each edge $(u,v)$ present in the dataset, we sample $\theta_{uv}$ uniformly at random in [0,0.2]. For each channel $u$, we generate a feature vector from that channel's row of the matrix, $\theta_{u}$ via complex nonlinear function. Specifically, $\theta_{u}$ is passed through a 5-layer neural network with random weight matrices and ReLU activations to obtain a feature vector $y_u$. The learning task is to reconstruct $\theta_{u}$ from $y_u$. The optimization task is to select $k$ channels in order to maximize the number of customers reached.
	
	\textbf{Bipartite matching: } This problem occurs in many domains; e.g., bipartite matching has been used to model the problem of a public housing programs matching housing resources to applicants \cite{benabbou2018diversity} or platforms matching advertisers with users \cite{bellur2007improved}. In each of these cases, the reward to matching any two nodes is not initially known, but is instead predicted from the features available for both parties. Bipartite matching can be formulated as a linear program, allowing us to apply our decision-focused approach. The learning problem is to use node features to predict whether each edge is present or absent (a classification problem). The optimization problem is to find a maximum matching in the predicted graph.
	
	Our experiments use the cora dataset \cite{sen2008collective}. The nodes are scientific papers and edges represent citation. Each node's feature vector indicating whether each word in a vocabulary appeared in the paper (there are 1433 such features). The overall graph has 2708 nodes. In order to construct instances for the decision problem, we partitioned the complete graph into 27 instances, each with 100 nodes, using metis \cite{karypis1998fast}. We divided the nodes in each instance into the sides of a bipartite graph (of 50 nodes each) such that the number of edges crossing sides was maximized. The learning problem is much more challenging than before: unlike in budget allocation, the features do not contain enough information to reconstruct the citation network. However, a decision maker may still benefit from leveraging whatever signal is available.
	
	\textbf{Diverse recommendation:}
	One application of submodular optimization is to select diverse sets of item, e.g.\ for recommendation systems or document summarization. Suppose that each item $i$ is associated with a set of topics $t(i)$. Then, we aim to select a set of $k$ items which collectively cover as many topics as possible: $f(S) = \left|\bigcup_{i \in S}t(i)\right|$.  Such formulations have been used across recommendation systems \cite{ashkan2015optimal}, text summarization \cite{takamura2009text}, web search \cite{agrawal2009diversifying} and image segmentation \cite{prasad2014submodular}.
	
	In many applications, the item-topic associations $t(i)$ are not known in advance. Hence, the learning task is to predict a binary matrix $\theta$ where $\theta_{ij}$ is 1 if item $i$ covers topic $j$ and 0 otherwise. The optimization task is to find a set of $k$ items maximizing the number of topics covered according to $\theta$. We consider a recommendation systems problem based on the Movielens dataset \cite{movielens} in which 2113 users rate 10197 movies (though not every user rated every movie). The items are the movies, while the topics are the top 500 actors. In our problem, the movie-actor assignments are unknown, and must be predicted only from user ratings. This is a \emph{multilabel classification problem} where we attempt to predict which actors are associated with each movie. We randomly divided the movies into 101 problem instances, each with 100 movies. The feature matrix $y$ contains the ratings given by each of the 2113 users for the 100 movies in the instance (with zeros where no rating is present).  
		\begin{table*}\centering
		%		\ra{1.3}
		\fontsize{9.5}{9.5}\selectfont
		\caption{Accuracy of each method according to standard measures.}\label{table:accuracy}
		\begin{tabular}{@{}r|ccccccc@{}}\toprule
			& \multicolumn{1}{c}{Budget allocation} & \phantom{abc} & \multicolumn{2}{c}{ Matching} & \phantom{abc} &\multicolumn{2}{c}{Diverse recommendation}\\
			\cmidrule{2-2} \cmidrule{4-5} \cmidrule{7-8} 
			& MSE && CE & AUC  && CE & AUC\\ \midrule
NN1-Decision & 0.8673e-02 $\pm$ 1.83e-04 && 0.994 $\pm$ 0.002 & 0.501 $\pm$ 0.011 && 1.053 $\pm$ 0.005 & 0.593 $\pm$ 0.003\\
NN2-Decision & 1.7118e-02 $\pm$ 2.65e-04 && 0.689 $\pm$ 0.004 & \textbf{0.560 $\pm$ 0.006} && 1.004 $\pm$ 0.022 & 0.577 $\pm$ 0.008\\
NN1-2Stage & 0.0501e-02 $\pm$ 2.67e-06 && 0.696 $\pm$ 0.001 & 0.499 $\pm$ 0.013 && 0.703 $\pm$ 0.001 & 0.389 $\pm$ 0.003\\
NN2-2Stage & 0.0530e-02 $\pm$ 2.27e-06 && \textbf{0.223 $\pm$ 0.005} & 0.498 $\pm$ 0.007 && 0.690 $\pm$ 0.000 & \textbf{0.674 $\pm$ 0.004}\\
RF-2Stage & \textbf{0.0354e-02 $\pm$ 4.17e-06} && 0.693 $\pm$ 0.000 & 0.500 $\pm$ 0.000 && \textbf{0.689 $\pm$ 0.000} & 0.500 $\pm$ 0.000\\
				\bottomrule
		\end{tabular}
		
	\end{table*}

	\textbf{Algorithms and experimental setup: } In each domain, we randomly divided the instances into 80\% training and 20\% test. All results are averaged over 30 random splits. Our decision-focused framework was instantiated using feed-forward, fully connected neural networks as the underlying predictive model. All networks used ReLU activations. We experimented with networks with 1 layer, representing a restricted class of models, and 2-layer networks, where the hidden layer (of size 200) gives additional expressive power. We compared two training methods. First, the decision-focused approach proposed above. Second, a two stage approach that uses a machine learning loss function (mean squared error for regression tasks and cross-entropy loss for classification). \emph{This allows us to isolate the impact of the training method since both use the same underlying architecture.} We experimented with additional layers but observed little benefit for either method. All networks were trained using Adam with learning rate $10^{-3}$. We refer to the 1-layer decision focused network as \emph{NN1-Decision} and the 1-layer two stage network as \emph{NN1-2Stage} (with analogous names for the 2-layer networks). We also compared to a random forest ensemble of 100 decisions trees (\emph{RF-2Stage}). Gradient-based training cannot be applied to random forests, so benchmark represents a strong predictive model which can be used by two stage approaches but not by our framework. Lastly, we show performance for a random decision. 
	
	\textbf{Solution quality: }
	Table \ref{table:opt} shows the solution quality that each approaches obtains on the full pipeline; i.e., the objective value of its decision evaluated using the true parameters. Each value is the mean (over the 30 iterations) and a bootstrapped 95\% confidence interval. For the budget allocation and diverse recommendation tasks, we varied the budget $k$. The decision-focused methods obtain the highest-performance across the board, tied with random forests on the synthetic budget allocation task. 
	
	We now consider each individual domain, starting with budget allocation. Both decision-focused methods substantially outperform the two-stage neural networks, obtaining at least 37\% greater objective value. This demonstrates that with fixed predictive architecture, decision-focused learning can greatly improve solution quality. NN1-Decision performs somewhat better than NN2-Decision, suggesting that the simpler class of models is easier to train. However, NN1-2Stage performs significantly worse than NN1-Decision, indicating that alignment between training and the decision problem is highly important for simple models to succeed. RF-2Stage performs essentially equivalently to NN1-Decision. This is potentially surprising since random forest are a much more expressive model class. As we will see later, much of the random forest's success is due to the fact that the features in this synthetic domain are very high-signal; indeed, they suffice for near-perfect reconstruction. The next two domains, both based on real data, explore low-signal settings where highly accurate recovery is  impossible.
	
	In bipartite matching, NN2-Decision obtains the highest overall performance, making nearly \emph{over 70\% more matches} than the next best method (RF-2Stage, followed closely by NN2-2Stage). Both 1-layer models perform extremely poorly, indicating that the more complex learning problem requires a more expressive model class. However, the highly expressive RF-2Stage does only marginally better than NN2-2Stage, demonstrating the critical role of aligning training and decision making.
	
	In the diverse recommendation domain, NN1-Decision has the best performance, followed closely by NN2-Decision. NN2-2Stage trails by 23\%, and NN1-2Stage performs extremely poorly. This highlights the importance of the training method within the same class of models: NN1-Decision obtains approximately 2.7 times greater objective value than NN1-2Stage. RF-2Stage also performs poorly in this domain, and is seemingly unable to extract any signal which boosts decision quality above that of random. 
	
	\begin{figure}
		\centering
		\includegraphics[width=3in]{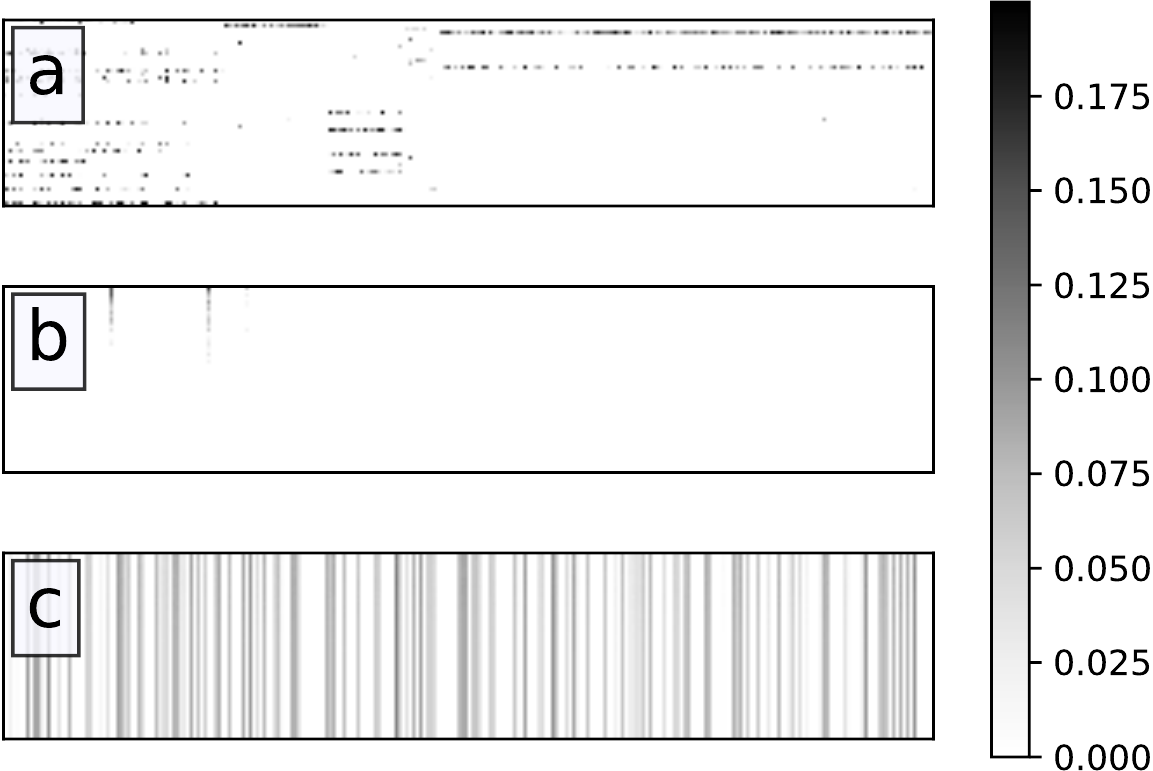}
		\caption{(a) ground truth (b) NN1-2Stage (c) NN1-Decision} \label{fig:visualization}
	\end{figure}

	\begin{figure} 
		\centering
		\includegraphics[width=1.5in]{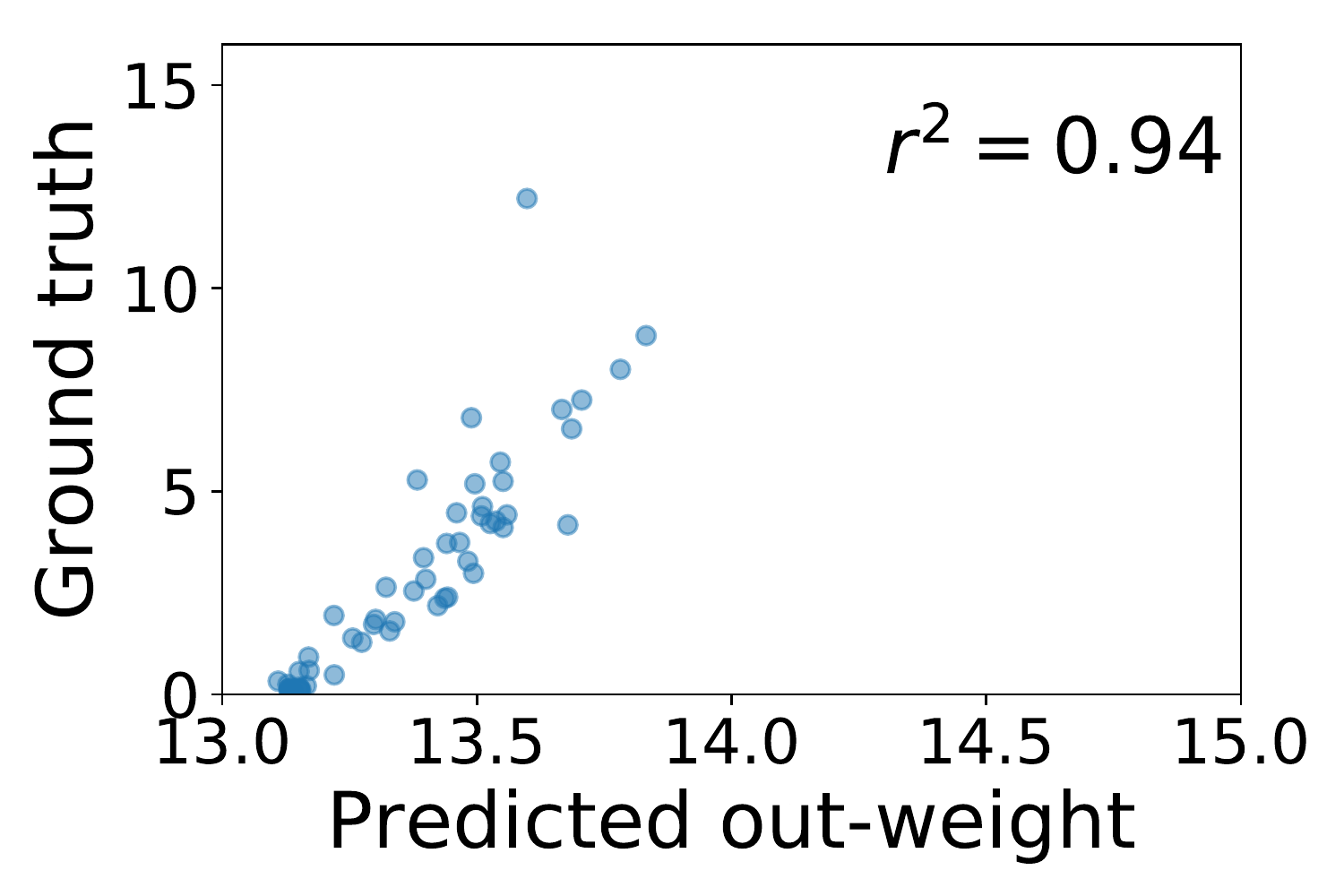}
		\includegraphics[width=1.5in]{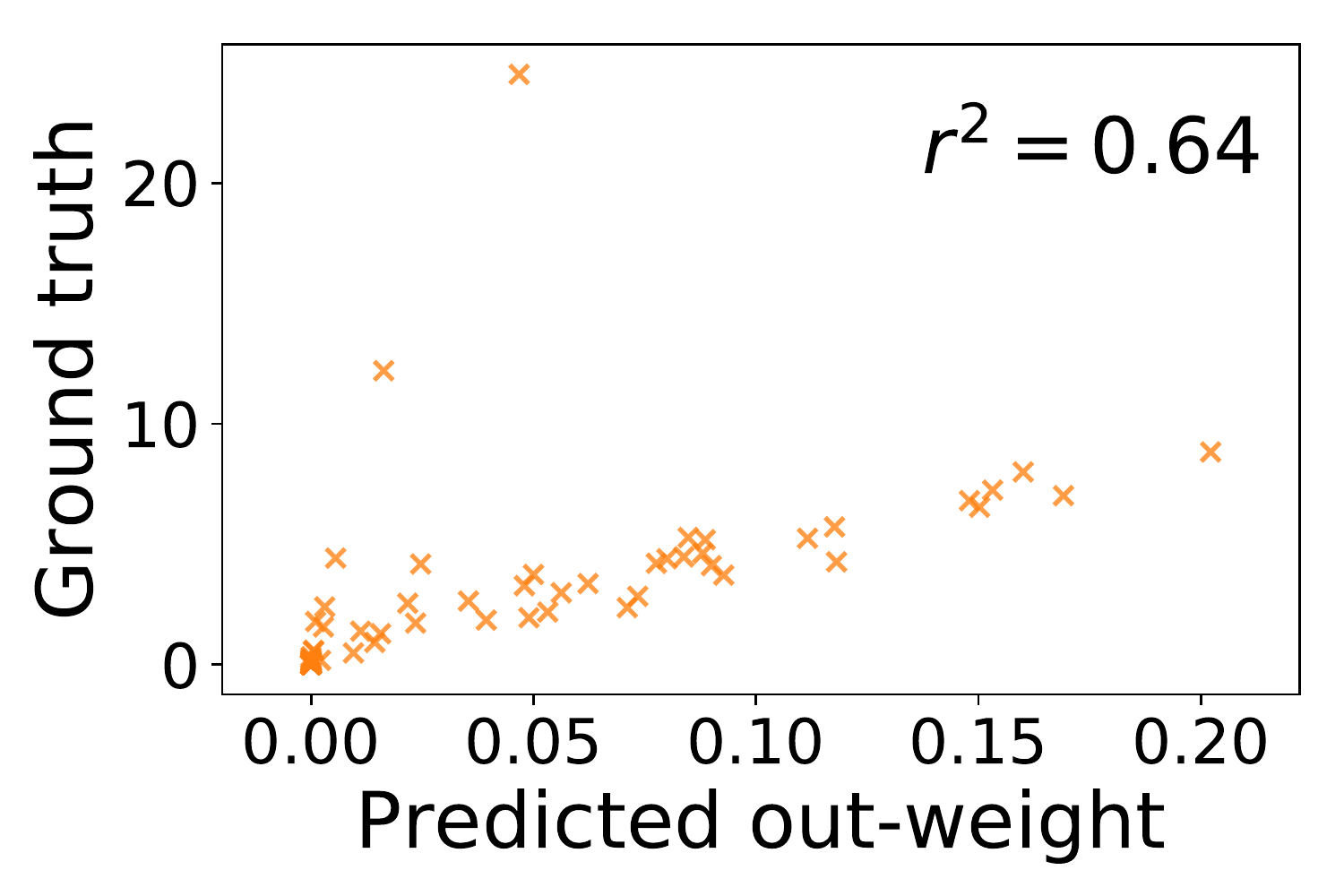}
		\caption{Left: our method's predicted total out-weight for each item. Right: predictions from two stage method.}\label{fig:outweight}
	\end{figure}
	\textbf{Exploration of learned models:}
	We start out by showing the accuracy of each method according to standard measures, summarized in Table \ref{table:accuracy}. For classification domains (diverse recommendation, matching), we show cross-entropy loss (which is directly optimized by the two stage networks) and AUC. For regression (the budget allocation domain), we show mean squared error (MSE). For budget allocation and diverse recommendation, we fixed $k=10$. 
	
	The two-stage methods are, in almost all cases, significantly more accurate than the decision-focused networks despite their worse solution quality. Moreoever, no accuracy measure is well-correlated with solution quality. On budget allocation, the two decision-focused networks have the worst MSE but the best solution quality. On bipartite matching, NN2-2Stage has better cross-entropy loss but much worse solution quality than NN2-Decision. On diverse recommendation, NN2-2Stage has the best AUC but worse solution quality than either decision-focused network.
	
	This incongruity raises the question of what differentiates the predictive models learned via decision-focused training. We now show more a more detailed exploration of each model's predictions. Due to space constraints, we focus on the simpler case of the synthetic budget allocation task, comparing NN1-Decision and NN1-2Stage. However, the higher-level insights generalize across domains (see the supplement for more detailed visualizations). 
	
	Figure \ref{fig:visualization} shows each model's predictions on an example instance. Each heat map shows a predicted matrix $\theta$, where dark entries correspond to a high prediction and light entries to low. The first matrix is the ground truth. The second matrix is the prediction made by NN1-2Stage, which matches the overall sparsity of the true $\theta$ but fails to recover almost all of the true connections. The last matrix corresponds to NN1-Decision and appears completely dissimilar to the ground truth. Nevertheless, these seemingly nonsensical predictions lead to the best quality decisions. 
	
	To investigate the connection between predictions and decision, Figure \ref{fig:outweight} aggregates each model's predictions at the channel level. Formally, we examine the predicted out-weight for each channel $u$, i.e., the sum of the row $\theta_u$. This is a coarse measure of $u$'s importance for the optimization problem; channels with connections to many customers are more likely to be good candidates for the optimal set. Surprisingly, NN1-Decision's predicted out-weights are extremely well correlated with the ground truth out-weights ($r^2 = 0.94$). However, the absolute magnitude of its predictions are skewed: the bulk of channels have low outweight (less than 1), but NN1-Decision's predictions are all at least 13. By contrast NN1-2Stage has poorer correlation, making it less useful for identifying the outliers which comprise the optimal set. However, it better matches the values of low out-weight channels and hence attains better MSE. This illustrates how aligning the model's training with the optimization problem leads it to focus on qualities which are specifically important for decision making, even if this compromises accuracy elsewhere.  
	
%	\textbf{Conclusion: } We propose a means of integrating a broad family of combinatorial optimization problems into the training of machine learning models by differentiating through solutions to a continuous relaxation of the discrete problem. This process aligns predictions with the end goals of a decision maker. Experimental results show that decision-focused learning can substantially improve solution quality (measured in terms of final optimization performance) across a variety of domains. By contrast, standard machine learning loss functions often fail to prioritize the qualities required for successful decision making. These results demonstrate that true end-to-end training is an important component of building a data-decisions pipeline.  
	
	\textbf{Acknowledgments: } This work was supported by the Army Research Office (MURI W911NF1810208) and a National Science Foundation Graduate Research Fellowship. 
	
	\bibliographystyle{aaai}
	
	\bibliography{ref}
	
	\appendix
	\normalsize
	
		\section{Proofs}

		\begin{proof}[Proof of Theorem 1]
			We start with the case where all rows of $A$ are linearly independent. Here, the result follows easily from Theorem 1 of \cite{amos2017optnet} since the Hessian matrix is $\gamma I$ and hence guaranteed to be positive definite. 
			
			When $A$ has linearly dependent rows, we argue that these rows can be removed without changing the feasible region. Consider two rows $a_i$ and $a_j$ such that for all $x$, $a_i^\top x = c a_j^\top x$ for some scalar $c$. We are guaranteed that the problem is feasible, meaning that there exists an $x$ which satisfies both constraints simultaneously. For this $x$, we have $a_i^\top x = b_i$ and $a_j^\top x = b_j$. But since $a_i^\top x = c a_j^\top x$, we must have $b_i = c b_j$. Accordingly, constraint $i$ is satisfied if and only if constraint $j$ is satisfied, and so removing one of the constraints leaves the feasible set unchanged. Applying this argument inductively yields the theorem. 
		\end{proof}

		\begin{proof}[Proof of Theorem 2]
			Let $x_{max} = \argmax_{y \in conv(\mathcal{X})} ||y||^2$. We have that 
			\begin{align*}
			\theta^\top x(\theta) &= \max_y \left[\theta^\top y - \gamma ||y||^2\right] + ||x(\theta)||^2\\
			&\geq \max_y \left[\theta^\top y\right] - \gamma ||x_{max}||^2 + \gamma ||x(\theta)||^2\\
			&= \max_y \left[\theta^\top y\right] + \gamma\left(||x(\theta)||^2 - ||x_{max}||^2\right)\\
			&\geq OPT - \gamma||x(\theta) - x_{max}||^2\\
			&\geq OPT - \gamma D
			\end{align*}
			
			where the second inequality uses the reverse triangle inequality. 
		\end{proof}

		\begin{proof}[Proof of Theorem 3]
			Since $\mathcal{X} = \{x \in \{0,1\}^{|V|} : \sum_i x_i \leq k \}$, $conv(\mathcal{X})$ is described by the two inequality constraints $-Ix \leq 0$ and $1^\top x \leq k$. It is easy to see that the corresponding constraint matrix $A$ has full row rank. Even though $F$ is not concave, any stationary point $(x, \lambda)$ must satisfy the KKT conditions. By applying the implicit function theorem to differentiate these equations, we get the form
			\begin{align*}\label{eq:kkt}
			\begin{bmatrix}
			&\nabla^2_x F(x, \theta) & A^T\\
			& diag(\lambda) A & diag(Ax-b)\\
			\end{bmatrix}
			\begin{bmatrix}
			\frac{d x}{d\theta}\\
			\frac{d \lambda}{d\theta}
			\end{bmatrix}
			=
			\begin{bmatrix}
			\frac{d \nabla_x f(x, \theta)}{d\theta} \\
			0
			\end{bmatrix}
			\end{align*} 
			
			So long as the right hand side matrix is invertible almost everywhere, the implicit function theorem guarantees that $\frac{dx}{d\theta}$ exists in a neighborhood of $x$ and satisfies the above conditions. Note that at a local maximum, we have $\nabla_x^2 F(x, \theta) \succ 0$, implying that the Hessian matrix must be invertible. Accordingly, it is easy to show that the RHS matrix is nonsingular by applying the same logic as \cite{amos2017optnet} (Theorem 1).
		\end{proof}
		
		\section{Visualizations}
		
		We now show more detailed analysis of the predictions made by each model in the other two domains: diverse recommendation and bipartite matching. The general trends are similar to those observed in the main paper for budget allocation (although the results are somewhat messier for the real-data domains). We see that the decision-focused neural network makes apparently nonsensical predications. However, the out-weight that it predicts for each item is better correlated with the ground truth than for the two stage method.  
		\begin{figure}
			\centering
			\includegraphics[width=3in]{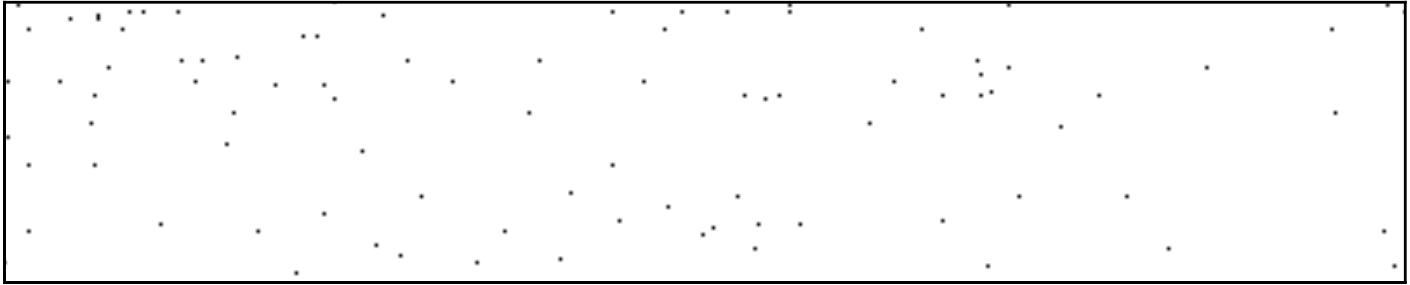}\\
			\includegraphics[width=3in]{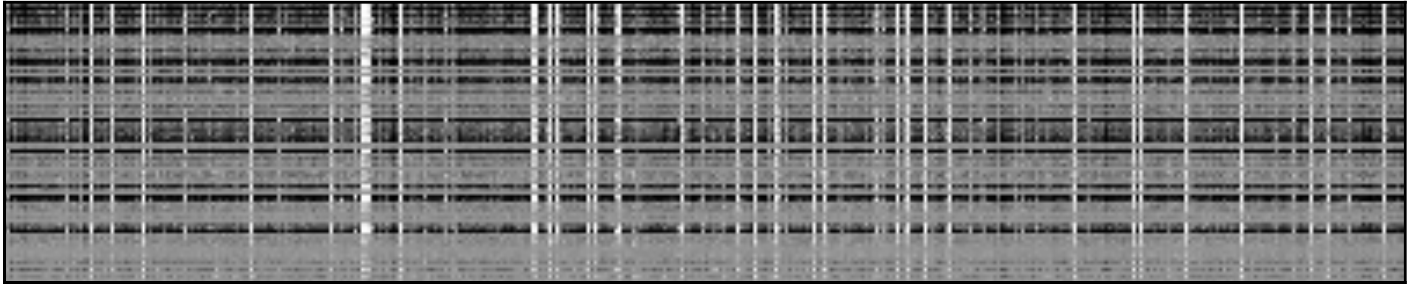}\\
			\includegraphics[width=3in]{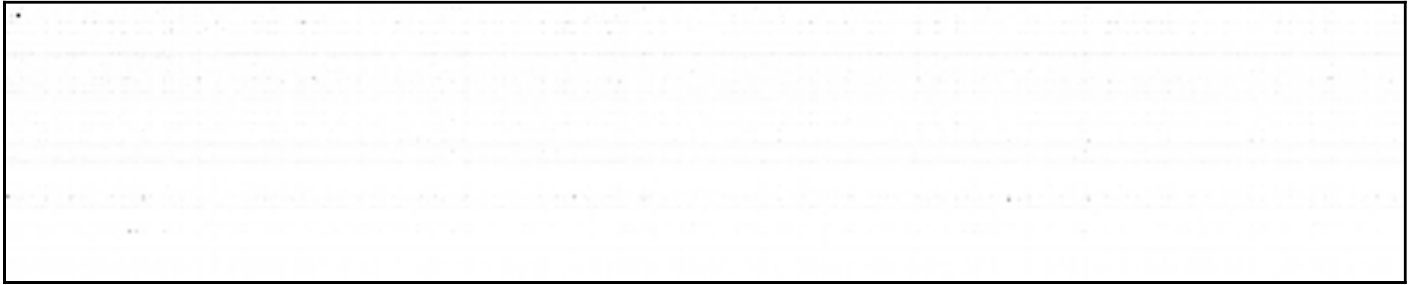}
			\caption{Diverse recommendation predictions. Top to bottom: ground truth, our method's prediction (by NN2-Decision), two stage prediction (by NN2-2Stage)}
		\end{figure}
		
		\begin{figure}
			\centering
			\includegraphics[width=1.5in]{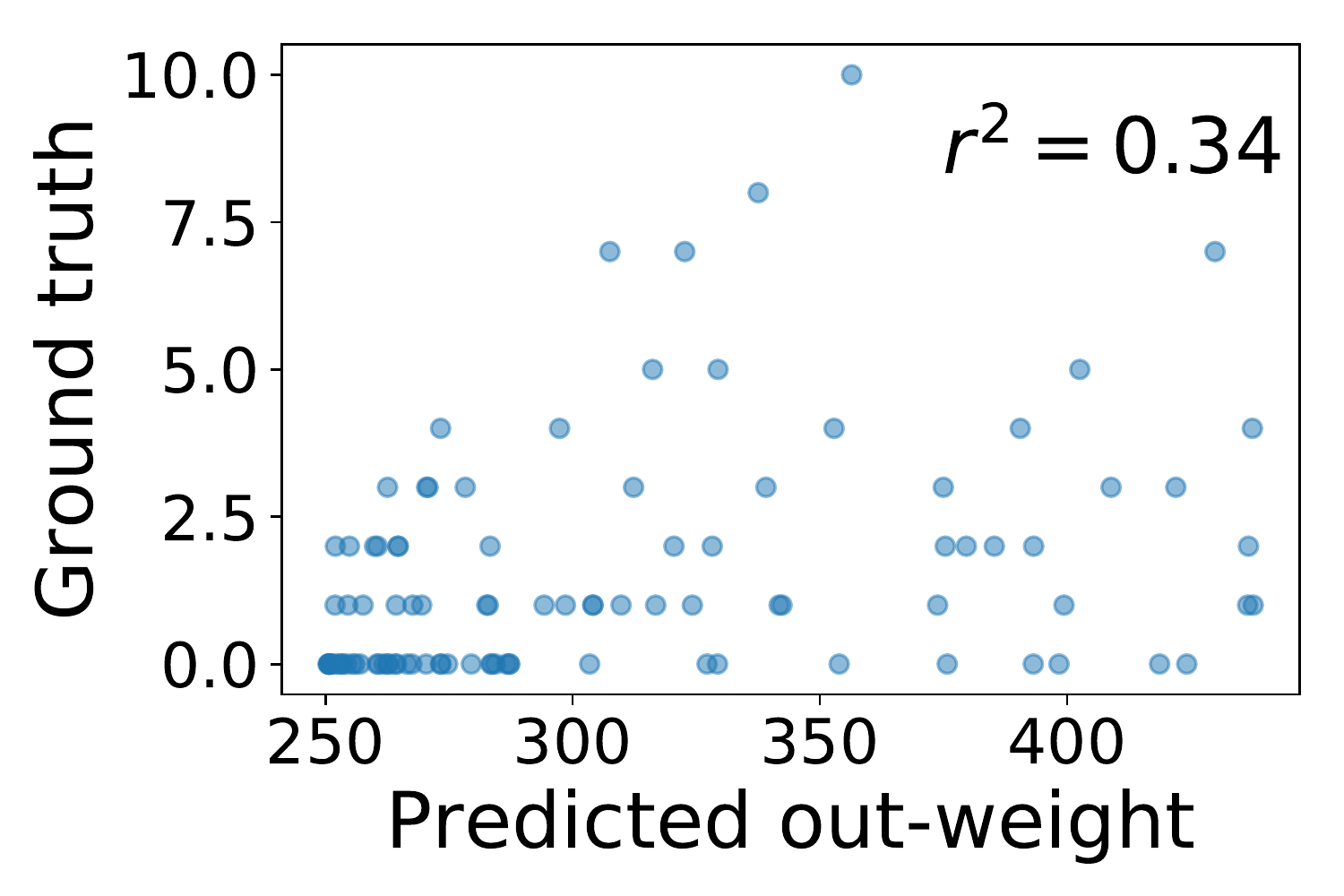}
			\includegraphics[width=1.5in]{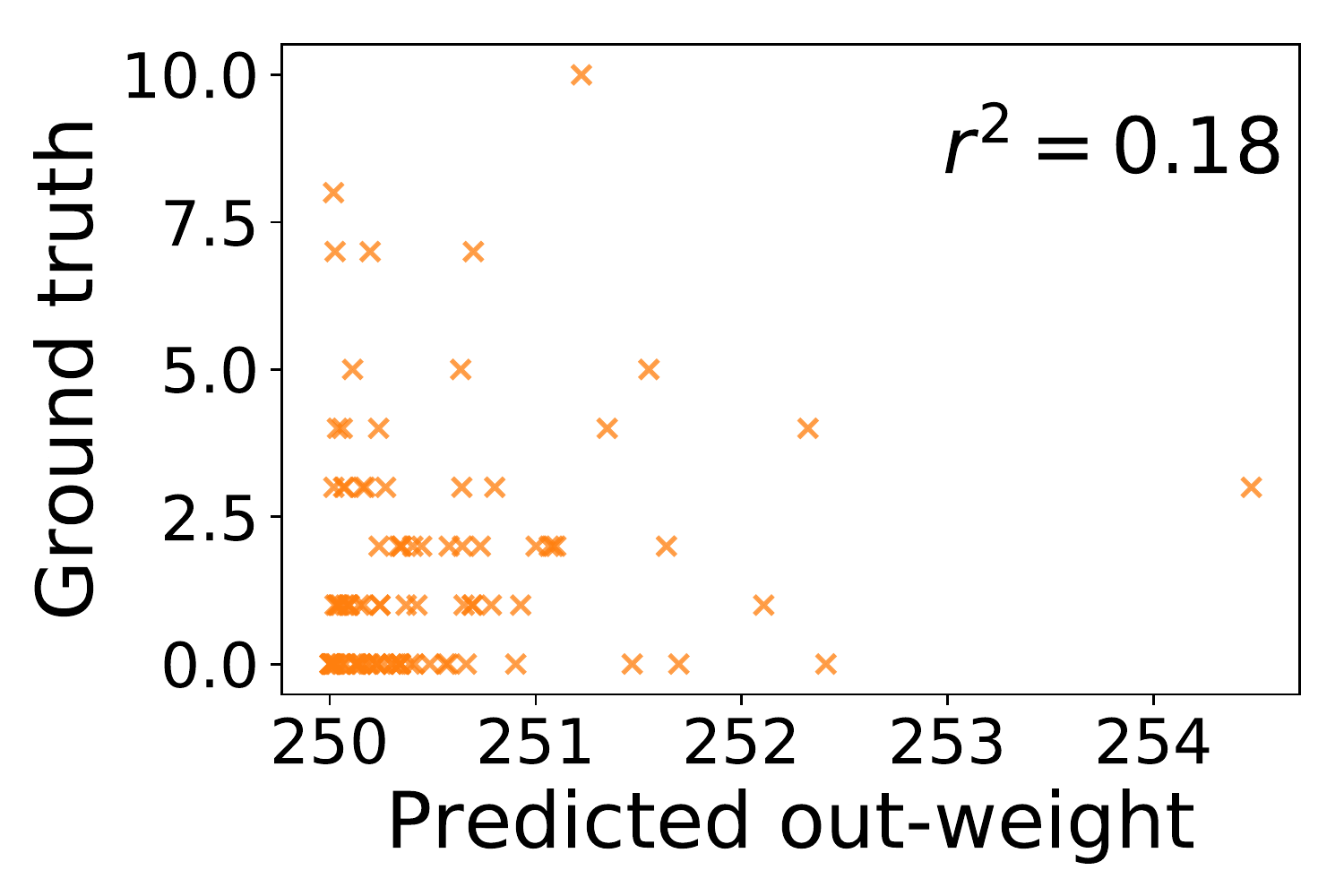}
			\caption{Diverse recommendation predicted outweight according to NN2-Decision (right) and NN2-2Stage (left).}
		\end{figure}
		
		\begin{figure}
			\centering
			\includegraphics[width=1.5in]{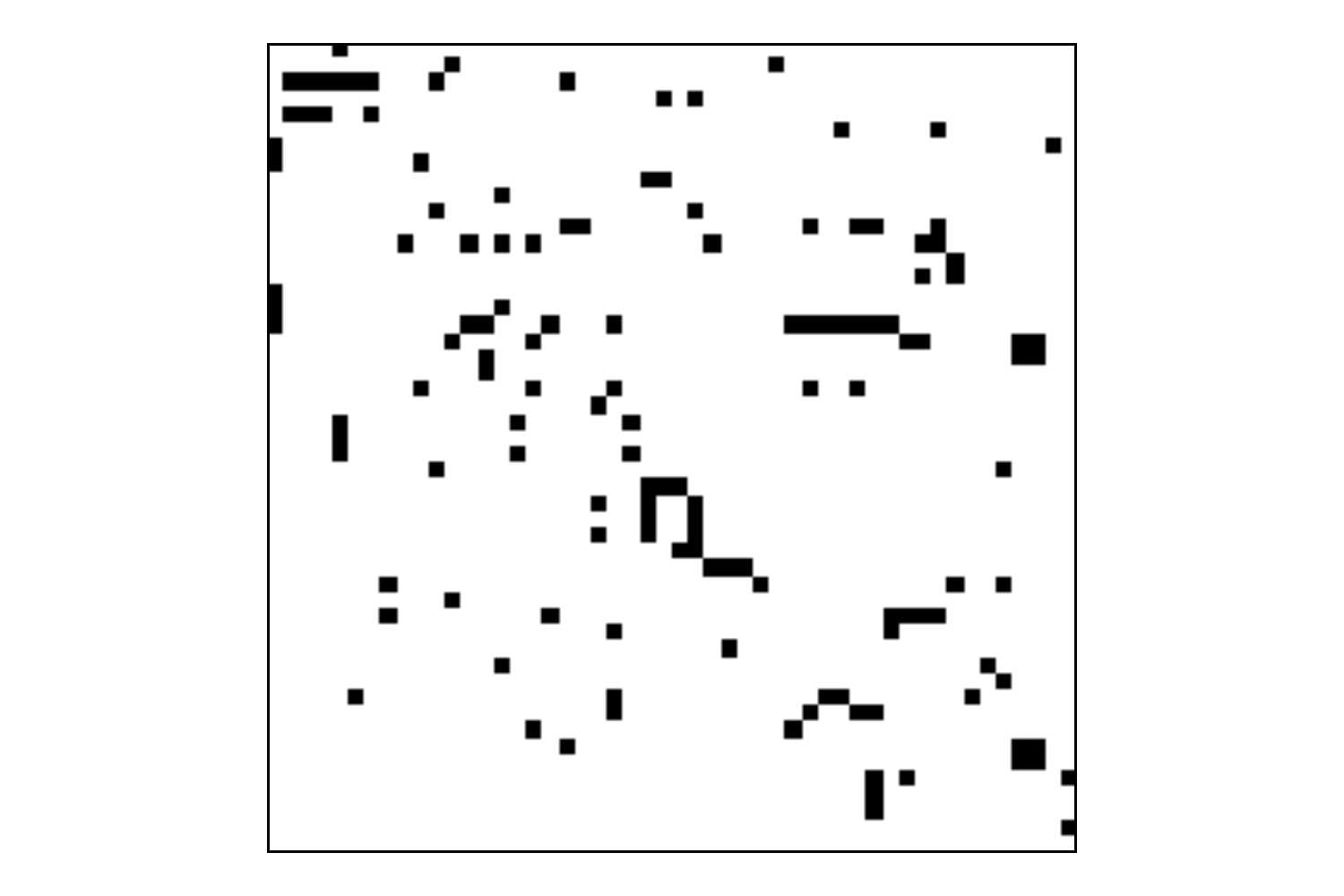}
			\includegraphics[width=1.5in]{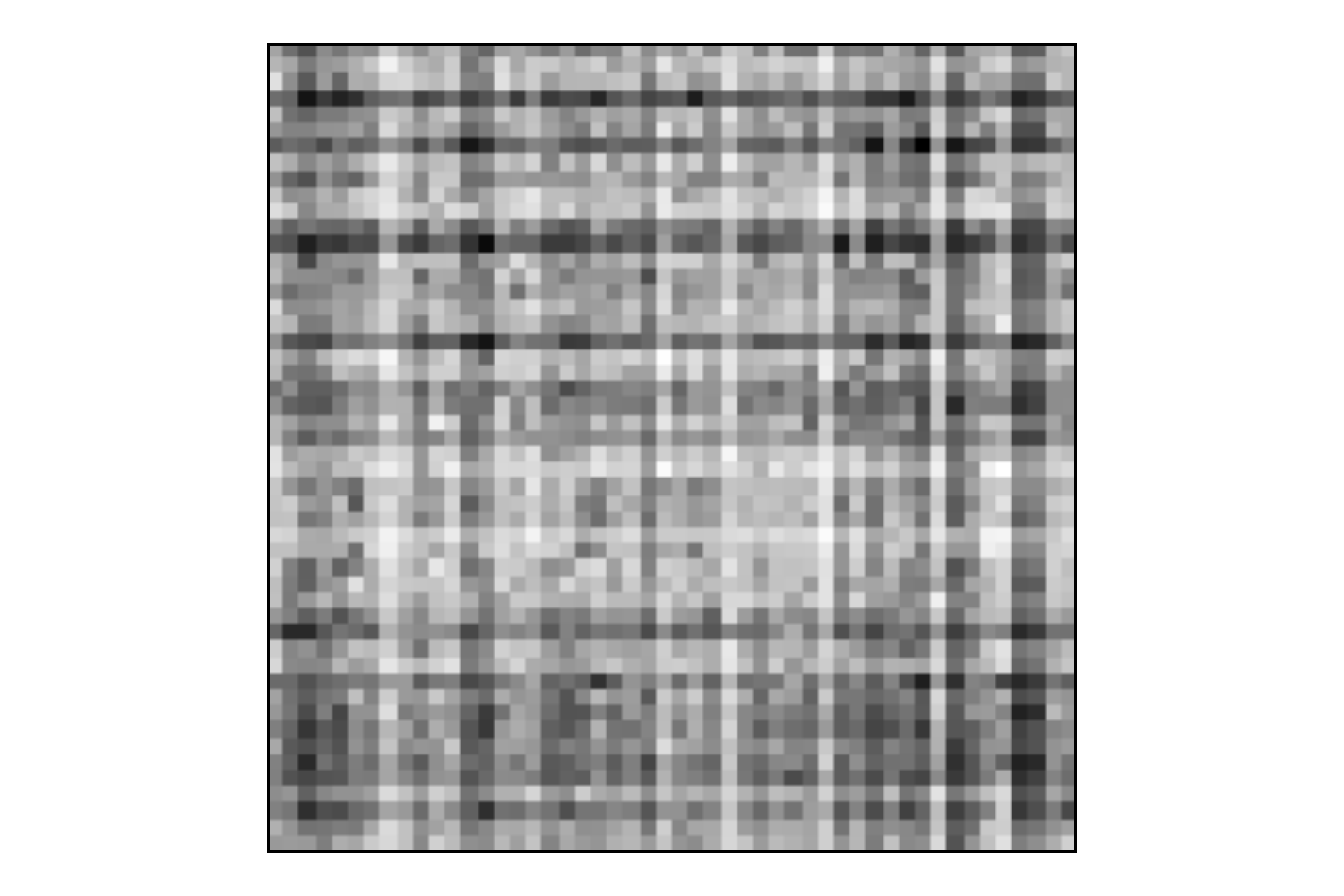}
			\includegraphics[width=1.5in]{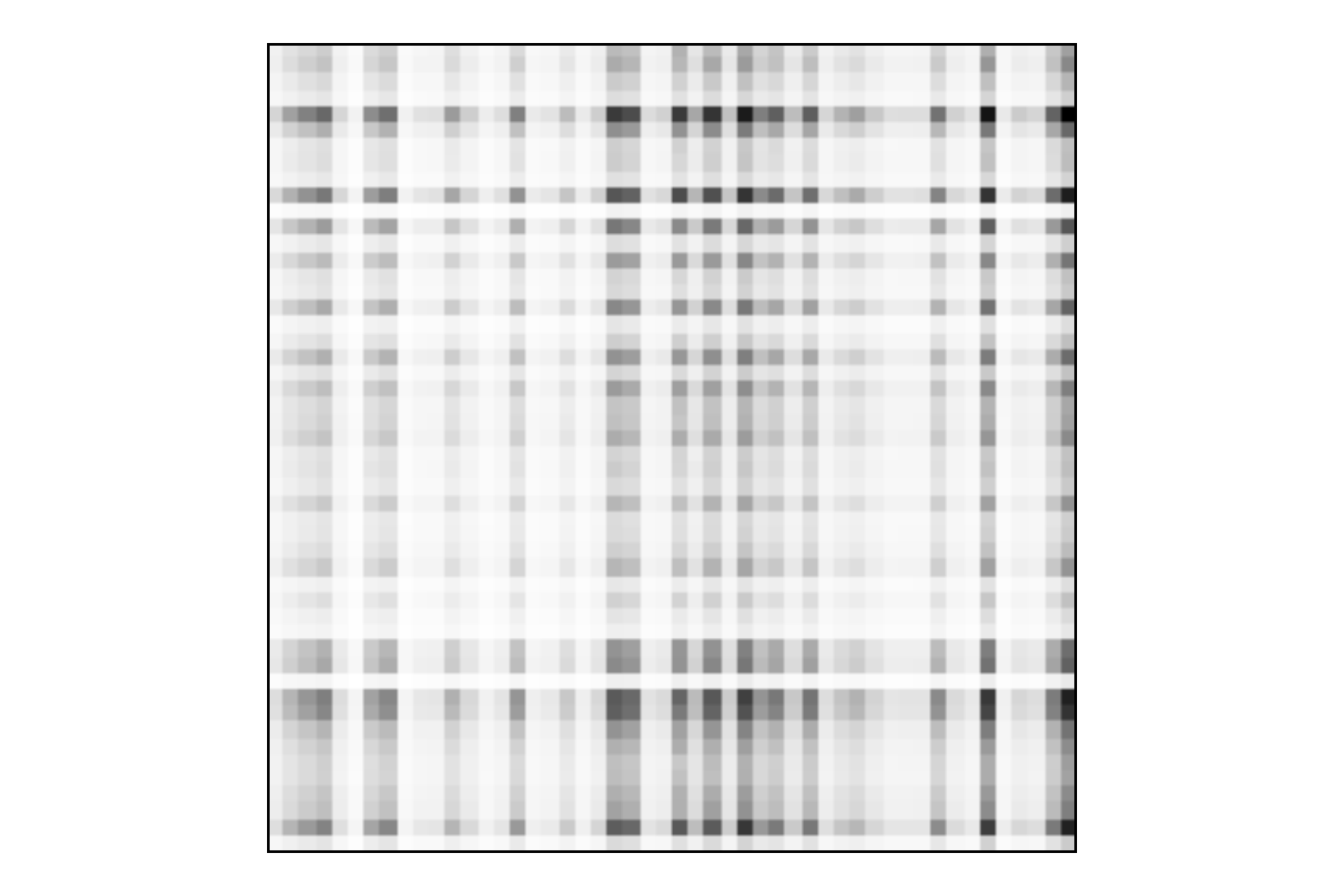}
			\caption{Bipartite matching predictions. Left to right: ground truth adjacency matrix, our method's prediction (NN2-Decision), two stage prediction (NN2-2Stage).}
		\end{figure}
		
		\begin{figure}
			\centering
			\includegraphics[width=1.5in]{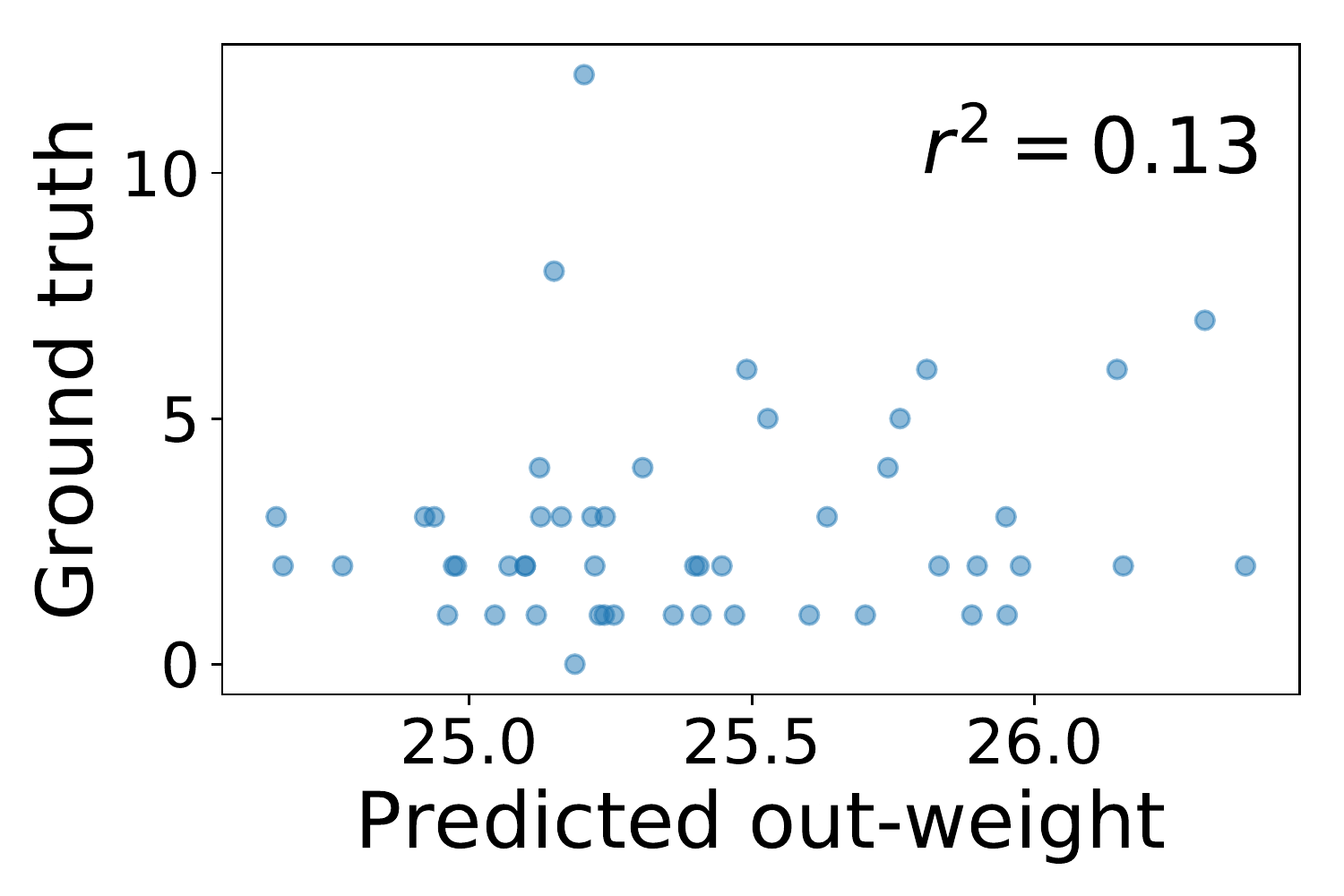}
			\includegraphics[width=1.5in]{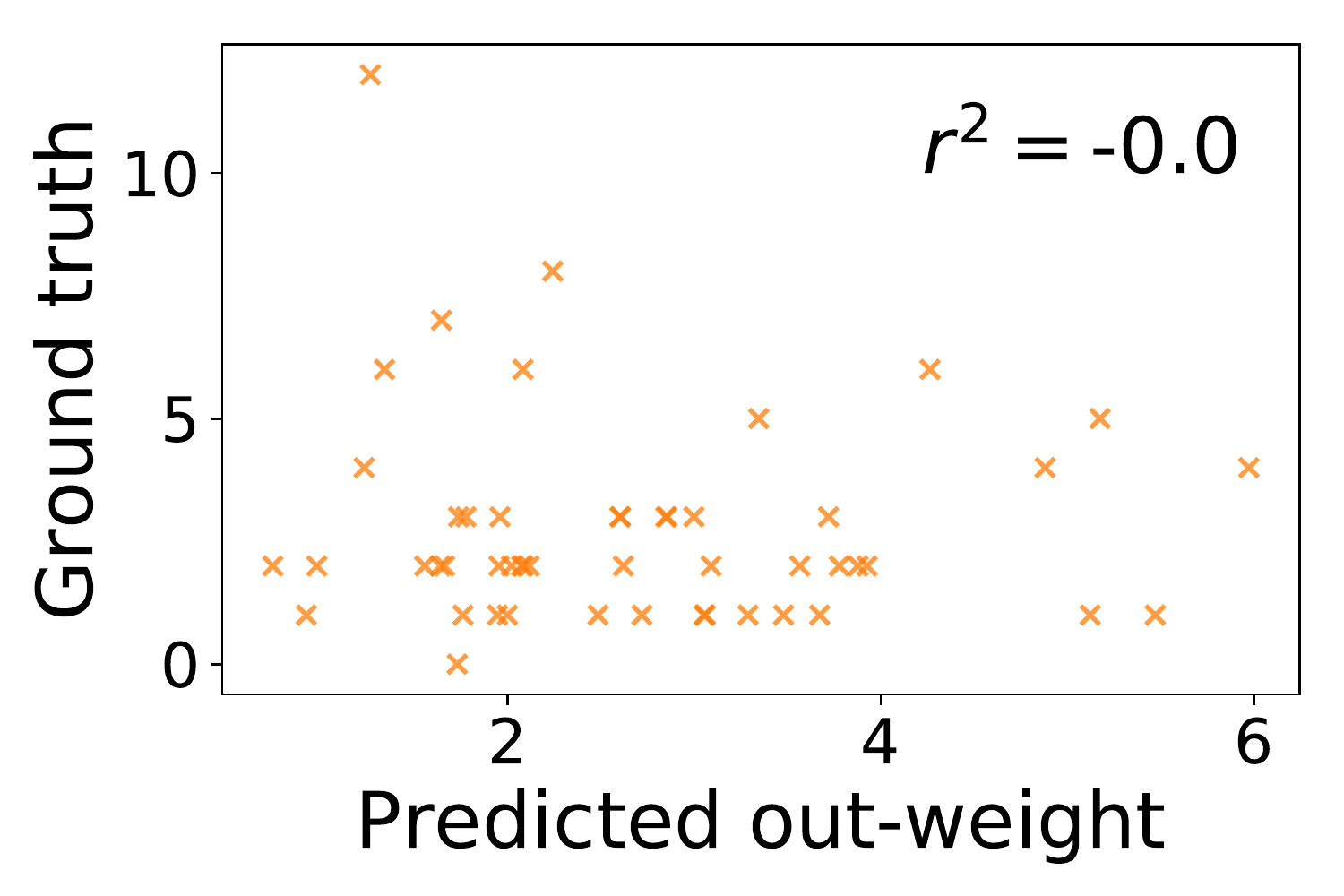}
			\caption{Bipartite matching predicted outweight according to NN2-Decision (right) and NN2-2Stage (left).}
		\end{figure}

\end{document}